\def\year{2021}\relax
\documentclass[letterpaper]{article} 
\usepackage{aaai21}  
\usepackage{times}  
\usepackage{helvet} 
\usepackage{courier}  
\usepackage[hyphens]{url}  
\usepackage{graphicx} 
\usepackage{natbib}  
\usepackage{caption} 
\usepackage{color}
\usepackage{overpic}
\usepackage{verbatim}
\usepackage{graphicx}
\usepackage{wrapfig}
\usepackage{amsmath}
\usepackage{booktabs}
\usepackage{diagbox}
\usepackage{multirow}
\usepackage{arydshln}
\usepackage[noend]{algpseudocode}
\usepackage{amssymb}
\usepackage[switch]{lineno}
\usepackage[dvipsnames]{xcolor}
\usepackage{algorithmicx,algorithm}
\title{Visual Boundary Knowledge Translation for Foreground Segmentation}
\author{
    Zunlei Feng\textsuperscript{\rm 1\#},
    Lechao Cheng\textsuperscript{\rm 2\#},
    Xinchao Wang\textsuperscript{\rm 3},
    Xiang Wang\textsuperscript{\rm 1},
    Yajie Liu\textsuperscript{\rm 2},
    Xiangtong Du\textsuperscript{\rm 4}, \\
    Mingli Song\textsuperscript{\rm 1}\thanks{Corresponding author. \textsuperscript{\#}Equal contribution to this work.}\\
}
\affiliations{
    \textsuperscript{\rm 1}Zhejiang University,
    \textsuperscript{\rm 2}Zhejiang Lab,
    \textsuperscript{\rm 3}Stevens Institute of Technology,
    \textsuperscript{\rm 4}Jiangsu University of Science and Technology
    \\

    \{zunleifeng, xiang.wang, brooksong\}@zju.edu.cn,
    \{chenglc,liuyj\}@zhejianglab.com, xinchao.wang@stevens.edu

}
\vspace{-2.0em}
\begin{document}

\def\mathbi#1{\textbf{\em #1}}

\maketitle

\begin{abstract}
\vspace{-0.2em}
When confronted with objects of unknown types in an image, humans can effortlessly and precisely tell their visual boundaries. This recognition mechanism and underlying generalization capability seem to contrast to state-of-the-art image segmentation networks that rely on large-scale {category-aware} annotated training samples. In this paper, we make an attempt towards building models that explicitly account for visual boundary knowledge, in hope to reduce the training effort on segmenting unseen categories.
Specifically, we investigate a new task  termed as Boundary Knowledge Translation~(BKT).
Given a set of fully labeled categories, BKT aims to \emph{translate} the visual boundary knowledge
learned from the labeled categories, to a set of novel categories, each of which is provided only a few labeled samples.
To this end, we propose a Translation Segmentation Network (Trans-Net), which comprises a segmentation network and two boundary discriminators. The segmentation network, combined with a boundary-aware self-supervised mechanism, is devised to conduct foreground segmentation, while the two discriminators work together in an adversarial manner to ensure an accurate segmentation of the novel categories under light supervision.
Exhaustive experiments demonstrate that, with only tens of labeled samples as guidance, Trans-Net achieves close results on par with fully supervised methods.
\end{abstract}

\vspace{-0.9em}
\section{Introduction}
Image segmentation has witnessed an
unprecedented development in the past decade
thanks to the deep learning.
The encouraging results, however, come at
the cost of the vast number of annotations
and  GPU training for days or even weeks.
To alleviate the training effort,
a number of learning techniques,
such as few-shot learning and transfer learning,
have proposed.
The former aims to train models using
only a few annotated samples,
while the later focuses on
transferring the models learned on one domain
to another novel one.
Despite the recent progress  in few-shot and transfer learning,
existing approaches are still prone to either inferior results~\citep{shaban2017one-shot,siam2019amp:,wang2019panet:},
or the {rigorous requirement that the two tasks are strongly related}~\citep{Dai2019Transfer,sun2019not}
and a large number of annotated samples~\citep{hong2017weakly,li2018weakly-,papandreou2015weakly-and}.

\begin{figure*}[!t]
\centering
\includegraphics[scale =0.90]{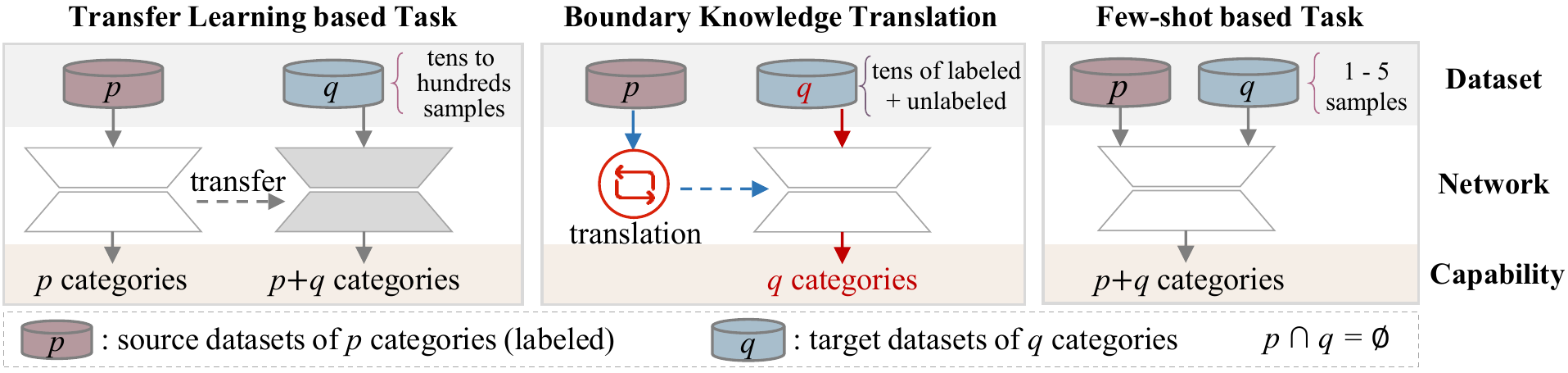}
\caption{
Comparing the proposed BKT with transfer learning and few-shot learning. Specifically, BKT \emph{translates} the knowledge learned from $p$ training categories into another $q$ ones, where the $p$ and $q$ categories are disjoint. Unlike transfer learning and few-shot learning based tasks where the trained models eventually tackle $p+q$ categories, {BKT  concentrates on the $q$ testing categories only.}
}
\vspace{-0.5em}
\label{fig:BKTvsOthers}
\end{figure*}

Nevertheless, our human eyes can effortlessly recognize the visual boundary of a scene object, even if this object seems unfamiliar or belongs to an unknown category.
Inspired by this fact, we study in this paper
a new Boundary Knowledge Translation Task (BKT-Task),
in aim to \emph{translate} the knowledge learned from
$p$ training categories where abundant annotations are available, into {another $q$ target categories} where a very small number of annotations are available for each class.

The differences between the proposed
BKT-Task and the two related tasks,
transfer learning and few-shot learning,
are summarized in Fig.~\ref{fig:BKTvsOthers}.
Unlike transfer learning that requires the $p$ source categories
and the $q$ target ones to be highly related so as to achieve reasonable performances,
in BKT such requirement is largely relaxed.
As demonstrated in our experiments, for example,
{the visual boundary knowledge of \emph{birds} can be
seamlessly translated into the segmentation network for \emph{flowers}.}
Also, in contrast to both transfer and few-shot learning,
whose outputs span $p+q$ categories,
BKT focuses on the $q$ categories only,
which seems to be a flaw of BKT but in reality not.
In many cases, since abundant annotations are available
for the $p$ categories, indicating that
state-of-the-art deep networks will highly likely to deliver
gratifying results,
a model that focuses on the novel $q$ categories
where annotations are insufficient is much desired.
Furthermore, as shown in prior works~\citep{chen2019unsupervised,minaee2020image,vinyals2016matching}
and also demonstrated in our experiments,
handling many categories simultaneously at once, especially
under a compact network architecture, will significantly
downgrade the performances for all classes;
focusing on the $q$ critical categories, on the other hand,
will alleviate this dilemma to a large extend
and ensure satisfactory performances.

To this end, we propose a Translation Segmentation Network (Trans-Net) for the above BKT-Task. The Trans-Net is designed to contain a segmentation network and two boundary discriminators. The segmentation network focuses on only segmenting the target categories, while the two discriminators collaborate in an
adversarial fashion: one works {on distinguishing whether the segmented foreground contains the background features,}
and the other {distinguishes whether the segmented background  contains the foreground features}.
Meanwhile, we also introduce pseudo masks {for enhancing and accelerating the translation of the boundary knowledge.}

For the segmentation network, we label tens of samples of
the target categories, as the guidance for segmenting the desired boundary of the foreground. We also propose
a self-supervised strategy  to
strengthen the boundary consistency of unlabeled samples.
Through the adversarial optimization, the visual boundary knowledge of fully labeled $p$ categories
can be effectively translated into the segmentation network by two boundary discriminators. Experiments demonstrate that, with only tens of labeled samples as guidance, Trans-Net achieves truly encouraging results on par with fully supervised methods.

Our contribution is therefore introducing
the new BKT-Task, in aim to
translate the visual boundary knowledge learned from
fully annotate source categories
into novel ones with few labels,
and proposing a dedicated solution,
Trans-Net, towards solving BKT-Task.
As the first attempt along its line,
Trans-Net, for the time being,  focuses
on foreground segmentation.
We devise a boundary-aware self-supervised mechanism and pseudo masks
for the segmentation network and the discriminators to enhance the translation of visual boundary knowledge, respectively.
We evaluate the proposed Trans-Net on a broad domain of image datasets, in term of both qualitative visualization and quantitative measures,
and show that the proposed method achieves results on par with
fully supervised ones.

\section{Related work}
We briefly review here two lines of work that are related to ours, GAN-based segmentation methods and boundary-aware segmentation.
More related works about \emph{few-shot learning} and \emph{transfer learning} are given in the supplementary materials.

\textbf{GAN-based segmentation methods} can be classified into two categories: mask distribution-based methods and composition fidelity based methods.
For the mask distribution-based methods, \citet{luc2016semantic} proposed the first GAN based semantic segmentation network, which adopts the adversarial optimization between segmented results and GT mask to train the segmentation network. Meanwhile, some researchers~\citep{arbelle2018microscopy,han2017transferring,xue2018segan:} applied the same adversarial strategy into medical image segmentation. With generated fake images and labeled images, \citet{souly2017semi} adopted the adversarial strategy to train the discriminator output class confidence maps. Furthermore, \citet{hung2018adversarial} extended the discriminator to generate a confidence map, which can be used to infer the regions sufficiently close to those from the ground truth distribution.
In the composition fidelity based methods~\citep{chen2019unsupervised,ostyakov2018seigan:,remez2018learning}, the segmented objects are firstly composited with some background images. Then, the discriminator is adopted to discriminate fidelity of the composited images and the GT nature images.
Unlike the above existing GAN-based methods, we adopt the adversarial strategy to translate the source categories' visual boundary knowledge into the segmentation network by two boundary discriminators.

\textbf{Boundary-aware segmentation.}
\citet{bertasius2016semantic} proposed two-tream framework where the predicted semantic boundaries is adopted to improve the semantic segmentation maps.
Similarly, two-tream framework is also adopted in some recent works~\citep{chen2016semantic,heng2017fusionNet:,yu2018learning,takikawa2019gated-scnn:},
where different constraints are devised for strengthening the segmentation results with the predicted boundaries.
Unlike predicting the boundary directly, ~\citet{hayder2017boundary-aware} proposed predicting pixels' distance to the object's boundary and post-processed the distance map into the final segmented results.
\citet{zhang2017global-residual} proposed a local boundary refinement network to learn the position-adaptive propagation coefficients so that local contextual information from neighbors can be optimally captured for refining object boundaries.
\citet{peng2017large} proposed a boundary refinement block to improve the localization performance near the object boundaries.
\citet{khoreva2017simple} proposed boundary-aware filtering to improve object delineation. \citet{qin2019basnet:} adopted the patch-level SSIM loss~\citep{wang2003multiscale} to assign higher weights to the boundary.
Unlike the above methods, we devise two boundary discriminators for discriminating the foreground's outer border and background's inner boundary, which can translate the visual boundary knowledge into a segmentation network for any new category.

\begin{figure*}[!t]
\centering
\includegraphics[scale =0.78]{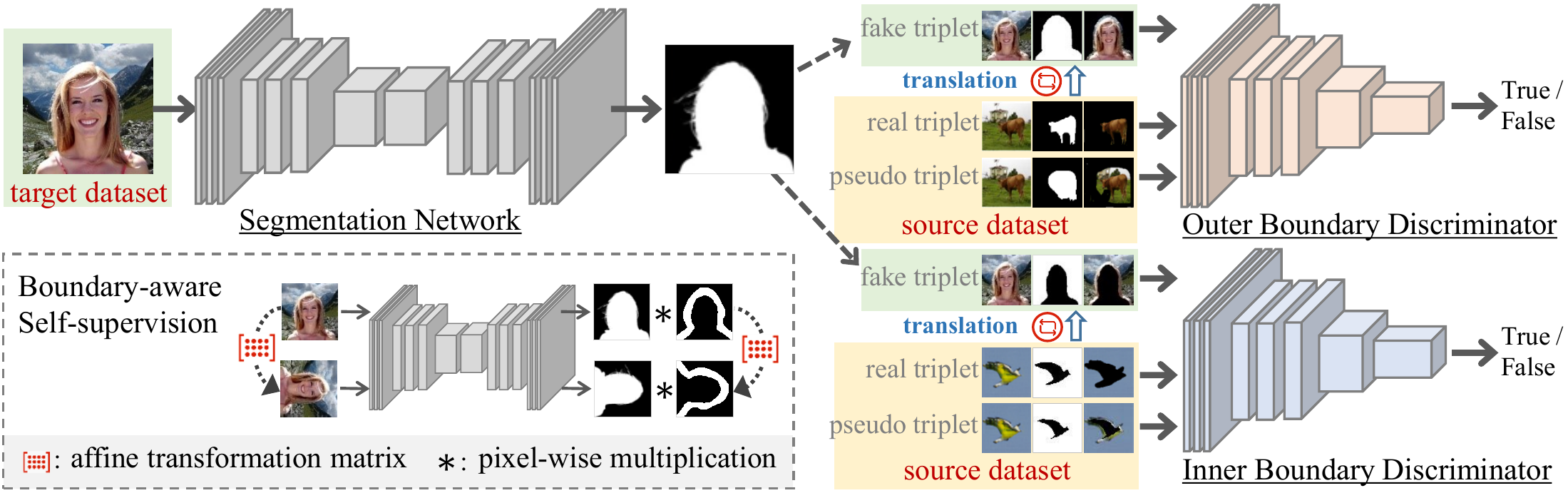}
\caption{The framework of Trans-Net. The segmentation network is designed to segment the sample's foreground in the target dataset. The outer boundary discriminator is devised for distinguishing whether the segmented foreground contains the outer background's features. The inner boundary discriminator is devised for distinguishing whether the segmented background contains the inner foreground's features. Pseudo samples of source dataset are generated with the eroded and dilated masks, which can reinforce the visual boundary knowledge translation. Boundary-aware self-supervision is proposed to constrain the boundary invariance on the target dataset.}
\label{fig:framework}
\vspace{-1em}
\end{figure*}

\section{Boundary Knowledge Translation Task}
The definition of the BKT-Task is given as follows. We assume that we are given the labeled source dataset $\mathbf{S}^p$ that contains $p$ object categories and target dataset $\mathbb{S}^q$ of $q$ object categories. The $p$ categories and $q$ categories are disjoint.
The visual boundary knowledge is defined as the perfect segmentation that the object doesn't contain outer background's features, and the background doesn't contain inner objects' features.
The goal of BKT-Task is to translate the visual boundary knowledge of $\mathbf{S}^p$ into the segmentation network $\mathcal{F}_{\theta}$, which is devised for only concentrating on the segmentation of $q$ categories.
The difference between the BKT-Task and other related tasks are shown in Fig.~\ref{fig:BKTvsOthers}.

\section{Method}
There are vast public labeled datasets for the segmentation task. The BKT-Task can effectively exploit those source datasets into the segmentation network for a new category, which will dramatically reduce the requirement for labeled samples in the new category.
As the first attempt along BKT-Task,
in this paper,
we focus on foreground segmentation and propose a Translation Segmentation Network (Trans-Net),  shown in Fig.~\ref{fig:framework}. Trans-Net contains a segmentation network and two boundary discriminators. The segmentation network is designed for only segment samples in the target dataset $\mathbb{S}^q$. The two boundary discriminators are devised for translating visual boundary knowledge of the source dataset $\mathbf{S}^p$ into the segmentation network without embezzling its capability for $q$ categories.

\subsection{Segmentation Network}

In  Trans-Net, the segmentation network $\mathcal{F}_{\theta}$ is designed to be an encoder-decoder architecture. Given a target image $\overline{\mathbi{x}}\in \mathbb{S}^q$, the segmented result $\tilde{\mathbi{m}}=\mathcal{F}_{\theta}(\overline{\mathbi{x}})$ is expected to approximate the GT mask $\mathbi{m}$, which can be achieved by minimizing the following basic reconstruction loss $\mathbf{\mathcal{L}_{rec}}$:
\begin{equation}\label{eq1}
\mathbf{\mathcal{L}_{rec}}=||\tilde{\mathbi{m}}-\mathbi{m}||^2_2,~\tilde{\mathbi{m}}=\mathcal{F}_{\theta}(\overline{\mathbi{x}}),~\overline{\mathbi{x}}\in \mathbb{S}^q.
\end{equation}
The $\mathbf{\mathcal{L}_{rec}}$ is used to reconstruct a few labeled samples of the target datasets, which will guide the segmentation of the desired boundary.

\emph{Boundary-aware Self-supervision.} To reduce the
amount of labelled samples,
inspired by \citet{wang2019self-supervised}, we propose a boundary-aware self-supervised strategy, which can strengthen the boundary consistency on target categories. The core idea is that the segmented result of the warped input image should be equal to the warped result of the input image. The schematic diagram of boundary-aware self-supervision is given in Fig.~\ref{fig:framework}. Formally, for the robust segmentation network, given an affine transformation matrix $\mathbi{A}$, segmented result $\mathcal{F}_{\theta}(\mathbi{A}\overline{\mathbi{x}})$ of the warped image $\mathbi{A}\overline{\mathbi{x}}$ and the warped result  $\mathbi{A}\mathcal{F}_{\theta}(\overline{\mathbi{x}})$ should be consistent in the following way: $\mathcal{F}_{\theta}(\mathbi{A}\overline{\mathbi{x}})=\mathbi{A}\mathcal{F}_{\theta}(\overline{\mathbi{x}})$. Furthermore, we obtain the boundary neighborhood weight map $\mathbi{w}$ as follows:
\begin{equation}\label{eq2}
\mathbi{w}=\mathfrak{D}_r(\tilde{\mathbi{m}})-\mathfrak{E}_r(\tilde{\mathbi{m}}),
\end{equation}
where, $\mathfrak{D}_r$ and $\mathfrak{E}_r$ denote the dilation and erosion operation with disk strel of radius $r$. The weight map $\mathbi{w}$ can further strengthen the boundary consistency. The boundary-aware self-supervised loss $\mathbf{\mathcal{L}_{sel}}$ is defined as follows:
\begin{equation}\label{eq3}
\mathbf{\mathcal{L}_{sel}}=||\mathbi{w}'\mathcal{F}_{\theta}(\mathbi{A}\overline{\mathbi{x}})-\mathbi{A}\{\mathbi{w}\mathcal{F}_{\theta}(\overline{\mathbi{x}})\}||^2_2,~\overline{\mathbi{x}}\in \mathbb{S}^q,
\end{equation}
where, $\mathbi{w}'$ and $\mathbi{w}$ are the weight maps of the predict masks $\mathcal{F}_{\theta}(\mathbi{A}\overline{\mathbi{x}})$ and $\mathcal{F}_{\theta}(\overline{\mathbi{x}})$, respectively. The boundary-aware self-supervised mechanism not only strengthens the boundary consistency but also can eliminate the unreasonable holes in the predict masks.

\subsection{Boundary Discriminator}
Inspired by the fact that humans can segment an object's boundary through distinguishing whether the inner and outer of boundary contain redundant features, we devise two boundary discriminators, which can translate the boundary knowledge of the source dataset $\mathbf{S}^p$ into the segmentation network.

\emph{Outer Boundary Discriminator.}
Given the input target image $\overline{\mathbi{x}}\in \mathbb{S}^q$, the segmentation network predicts the mask $\tilde{\mathbi{m}}=\mathcal{F}_{\theta}(\overline{\mathbi{x}})$. Next, the foreground $\overline{\mathbi{x}}^o$ is computed using the following equation: $\overline{\mathbi{x}}^o=\tilde{\mathbi{m}}*\overline{\mathbi{x}}$, where `$*$' denotes pixel-wise multiplication. Then the concatenated triplet $\mathbi{I}^o_a=[\overline{\mathbi{x}},\tilde{\mathbi{m}},\overline{\mathbi{x}}^o]$ is input into the outer boundary discriminator $\mathcal{D}^o_{\phi}$, which discriminates whether the segmented foreground $\overline{\mathbi{x}}^o$ contains the background's outer features. In the paper, $\mathbi{I}^o_a$ is regarded as a fake triplet.
Meanwhile, choosing a labeled sample $\underline{\mathbi{x}}$ from the source dataset $\mathbf{S}^p$, the corresponding $\mathbi{I}^o_e=[\underline{\mathbi{x}},\mathbi{m},\underline{\mathbi{x}}^o]$ is labeled as a real triplet.
Furthermore, we reprocess the GT mask $\mathbi{m}$ of samples $\underline{\mathbi{x}}\in \mathbf{S}^p$ by dilation operation and get the pseudo triplet $\mathbi{I}^o_s=[\underline{\mathbi{x}},\mathfrak{D}_r(\mathbi{m}),\underline{\mathbi{x}}^o_{\mathfrak{D}}]$, where $ \underline{\mathbi{x}}\in \mathbf{S}^p$ and $\underline{\mathbi{x}}^o_{\mathfrak{D}}=\mathfrak{D}_r(\mathbi{m})*\underline{\mathbi{x}}$.
The generated pseudo triplet $\mathbi{I}^o_s$ will assist the outer boundary discriminator in distinguishing the background's outer features.
The adversarial optimization between the segmentation network and outer boundary discriminator will translate the outer boundary knowledge of the source dataset $\mathbf{S}^p$ into the segmentation network with the following outer boundary adversarial loss $\mathbf{\mathcal{L}^{out}_{adv}}$:
\begin{equation}\label{eq4}
\begin{split}
\mathbf{\mathcal{L}^{out}_{adv}}\!\!= & \frac{1}{2}\!\mathop{\mathbb{E}}\limits_{\mathbi{I}^o_a \sim \mathbb{P}^o_a}\![\mathcal{D}^o_{\phi}\!(\mathbi{I}^o_a)]\!+\!
\frac{1}{2}\!\mathop{\mathbb{E}}\limits_{\mathbi{I}^o_s \sim \mathbb{P}^o_s} [\mathcal{D}^{o}_{\phi}\!(\mathbi{I}^o_s)]\! -\!\!\!\!\mathop{\mathbb{E}}\limits_{\mathbi{I}^o_e \sim \mathbb{P}^o_e}\![\mathcal{D}^{o}_{\phi}\!(\mathbi{I}^o_e)] \\
& +\lambda \! \mathop{\mathbb{E}}\limits_{\mathbi{I}^o\sim \mathbb{P}_{\mathbi{I}^o}}\!\![(\|\nabla_{\mathbi{I}^o} \mathcal{D}^{o}_{\phi}(\mathbi{I}^o)\|_{2}\!-\!1)^2],
\end{split}
\end{equation}
where, the $ \mathbb{P}^o_a$, $\mathbb{P}^o_s$, $\mathbb{P}^o_e$ are the segmented outer boundary distribution, pseudo outer boundary distribution, and real outer boundary distribution, respectively.
The $\mathbb{P}_{\mathbi{I}^o}$ is sampled uniformly along straight lines between pairs of points sampled from the distribution $\mathbb{P}^o_e$ and the segmentation network distribution $ \mathbb{P}^o_a$.
The $\mathbi{I}^o=\varepsilon \mathbi{I}^o_e+(1-\varepsilon)\mathbi{I}^o_a$, where the $\varepsilon$ is a random number between $0$ and $1$.
The gradient penalty term is firstly proposed in WGAN-GP~\cite{gulrajani2017improved}. The $\lambda$ is the gradient penalty coefficient.

\begin{algorithm}[!th]
\caption{The Training Algorithm for Trans-Net}
\begin{algorithmic}[1]
\renewcommand{\algorithmicrequire}{\textbf{Require:}}
\renewcommand{\algorithmicensure}{\textbf{Require:}}
\Require{ The gradient penalty coefficient $\lambda$, interval iteration number
$n_{critic}$, the batch size $K$, Adam hyperparameters $\alpha,\beta_{1},\beta_{2}$, the balance parameters $\tau,\eta$ for $\mathcal{L}_{rec}$ and $\mathbf{\mathcal{L}_{sel}}$, the Laplace smoothing parameter $\xi$.}
\Require{ Initial critic parameters $\varphi,\phi$, initial segmentation network parameters $\theta$. }
\While{$\theta$ has not converged}
\For{$t=1,...,n_{critic}$}
\For{$k=1,...,K$}
\State Sample $\overline{\mathbi{x}}$ from target dataset $\mathbb{S}^q$, $(\underline{\mathbi{x}},\mathbi{m})$ from source dataset $\mathbf{S}^p$, a random number $\varepsilon \sim U[0,1]$.
\State Obtain real triplet $\mathbi{I}^o_e=[\underline{\mathbi{x}},\mathbi{m},\underline{\mathbi{x}}^o]$ and $\mathbi{I}^i_e=[\underline{\mathbi{x}},\mathbi{m}',\underline{\mathbi{x}}^i]$, $\underline{\mathbi{x}}^o=\mathbi{m}*\underline{\mathbi{x}}$, $\underline{\mathbi{x}}^i=\mathbi{m}'*\underline{\mathbi{x}}$.
\State Obtain fake triplet $\mathbi{I}^o_a=[\overline{\mathbi{x}},\tilde{\mathbi{m}},\overline{\mathbi{x}}^o]$ and $\mathbi{I}^i_a=[\overline{\mathbi{x}},\tilde{\mathbi{m}}',\overline{\mathbi{x}}^i]$, $\tilde{\mathbi{m}}=\mathcal{F}_{\theta}(\overline{\mathbi{x}})$, $\tilde{\mathbi{m}}'=[\mathbf{1}]-\tilde{\mathbi{m}}$.
\State Obtain pseudo triplet $\mathbi{I}^o_s=[\underline{\mathbi{x}},\mathfrak{D}_r(\mathbi{m}),\underline{\mathbi{x}}^o_{\mathfrak{D}}]$ and $\mathbi{I}^i_s=[\underline{\mathbi{x}},\mathfrak{E}_r(\mathbi{m}'),\underline{\mathbi{x}}^i_{\mathfrak{E}}]$.
\State $\mathbi{I}^o \leftarrow  \varepsilon \mathbi{I}^o_e+(1-\varepsilon)\mathbi{I}^o_a$, $\mathbi{I}^i \leftarrow  \varepsilon \mathbi{I}^i_e+(1-\varepsilon)\mathbi{I}^i_a$.
\State $\mathbf{\mathcal{L}^{out}_{adv}}^{(k)} \leftarrow \frac{1}{2}\mathcal{D}^o_{\phi}(\mathbi{I}^o_a)  + \frac{1}{2}\mathcal{D}^{o}_{\phi}(\mathbi{I}^o_s) - \mathcal{D}^{o}_{\phi}(\mathbi{I}^o_e) +$ \Statex \quad \qquad \qquad \qquad \qquad $\lambda (\|\nabla_{\mathbi{I}^o} \mathcal{D}^{o}_{\phi}(\mathbi{I}^o)\|_{2}-1)^2$.
\State $\mathbf{\mathcal{L}^{in}_{adv}}^{(k)} \leftarrow \frac{1}{2}\mathcal{D}^i_{\varphi}(\mathbi{I}^i_a)  + \frac{1}{2}\mathcal{D}^{i}_{\varphi}(\mathbi{I}^i_s) - \mathcal{D}^{i}_{\varphi}(\mathbi{I}^i_e) + $ \Statex \quad \qquad \qquad \qquad \qquad$\lambda (\|\nabla_{\mathbi{I}^i} \mathcal{D}^{i}_{\varphi}(\mathbi{I}^i)\|_{2}-1)^2$.
\EndFor
\State $\phi \leftarrow Adam(\nabla_{\phi}\frac{1}{K} \sum^K_{k=1}\mathbf{\mathcal{L}^{out}_{adv}}^{(k)},\phi,\alpha,\beta_1,\beta_2)$.
\State $\varphi \leftarrow Adam(\nabla_{\varphi}\frac{1}{K} \sum^K_{k=1}\mathbf{\mathcal{L}^{in}_{adv}}^{(k)},\varphi,\alpha,\beta_1,\beta_2)$.
\EndFor
\State Sample unlabeled batch $\{\overline{\mathbi{x}}^{(k)}\}^K_{k=1}$ and labeled batch $\{\ddot{\overline{\mathbi{x}}}^{(k)},\mathbi{m}^{(k)}\}^K_{k=1}$ from target dataset $\mathbb{S}^q$.
\State $\mathbf{\mathcal{L}_{rec}}\leftarrow ||\mathcal{F}_{\theta}(\ddot{\overline{\mathbi{x}}})-\mathbi{m}||^2_2$.
\State  $\mathbf{\mathcal{L}_{sel}}\leftarrow ||\mathbi{w}'\mathcal{F}_{\theta}(\mathbi{A}\overline{\mathbi{x}})-\mathbi{A}\{\mathbi{w}\mathcal{F}_{\theta}(\overline{\mathbi{x}})\}||^2_2$.
\State $\theta \leftarrow Adam(\nabla_{\theta}\frac{1}{K} \sum^K_{k=1}\{\mathbf{\tau\mathcal{L}_{rec}}+ \eta\mathbf{\mathcal{L}_{sel}} -\mathcal{D}^{o}_{\phi}(\mathcal{F}_{\theta}(\overline{\mathbi{x}}))-\mathcal{D}^{i}_{\varphi}(\mathcal{F}_{\theta}(\overline{\mathbi{x}}))\}),\theta,\alpha,\beta_1,\beta_2)$.
\EndWhile
\State \Return Segmentation network parameters $\theta$, critic parameters $\varphi,~\phi$.
\end{algorithmic}
\label{alg:alg1}
\end{algorithm}

\emph{Inner Boundary Discriminator.}
The inner boundary discriminator
$\mathcal{D}^i_{\varphi}$ is devised for discriminating whether the segmented
background contains the object's inner features.
To obtain the segmented background, the predict background mask $\tilde{\mathbi{m}}'$ and GT mask $\mathbi{m}'$
are reprocessed with the Not-operation as follows: $\tilde{\mathbi{m}}'=[\mathbf{1}]-\tilde{\mathbi{m}}$,
$\mathbi{m}'=[\mathbf{1}]-\mathbi{m}$, where the $[\mathbf{1}]$
denotes the unit matrix of $\mathbi{m}$'s size.
Then, the corresponding fake triplet $\mathbi{I}^i_a=[\overline{\mathbi{x}},\tilde{\mathbi{m}}',\overline{\mathbi{x}}^i]$, real triplet $\mathbi{I}^i_e=[\underline{\mathbi{x}},\mathbi{m}',\underline{\mathbi{x}}^i]$ and pseudo triplet $\mathbi{I}^i_s=[\underline{\mathbi{x}},\mathfrak{D}_r(\mathbi{m}'),\underline{\mathbi{x}}^i_{\mathfrak{D}}]$ are computed in the same manner as done in the outer boundary discriminator.
The generated pseudo triplet $\mathbi{I}^i_s$ will also assist the inner boundary discriminator in distinguishing the inner features of the foreground. Similarly, the inner boundary adversarial loss $\mathbf{\mathcal{L}^{in}_{adv}}$ is defined as follows:
\begin{equation}\label{eq5}
\begin{split}
\mathbf{\mathcal{L}^{in}_{adv}}\!\!= & \frac{1}{2}\!\mathop{\mathbb{E}}\limits_{\mathbi{I}^i_a \sim \mathbb{P}^i_a}\![\mathcal{D}^i_{\varphi}\!(\mathbi{I}^i_a)]\!+\!
\frac{1}{2}\!\mathop{\mathbb{E}}\limits_{\mathbi{I}^i_s \sim \mathbb{P}^i_s} [\mathcal{D}^i_{\varphi}\!(\mathbi{I}^i_s)]\!-\!\!\!\!\mathop{\mathbb{E}}\limits_{\mathbi{I}^i_e \sim \mathbb{P}^i_e}\![\mathcal{D}^i_{\varphi}\!(\mathbi{I}^i_e)]\\
& + \lambda \!\mathop{\mathbb{E}}\limits_{\mathbi{I}^i\sim \mathbb{P}_{\mathbi{I}^i}}\!\![(\|\nabla_{\mathbi{I}^i} \mathcal{D}^i_{\varphi}(\mathbi{I}^i)\|_{2}\!-\!1)^2],
\end{split}
\end{equation}
where, the $ \mathbb{P}^i_a$, $\mathbb{P}^i_s$, $\mathbb{P}^i_e$ are the segmented inner boundary distribution, pseudo inner boundary distribution, and real inner boundary distribution, respectively.
$\mathbi{I}^i=\varepsilon \mathbi{I}^i_e+(1-\varepsilon)\mathbi{I}^i_a$. The optimization on $\mathbf{\mathcal{L}^{in}_{adv}}$ will translate the inner boundary knowledge of the source dataset $\mathbf{S}^p$ into the segmentation network.

\begin{figure*}[!t]
\centering
\begin{overpic}[scale =0.60]{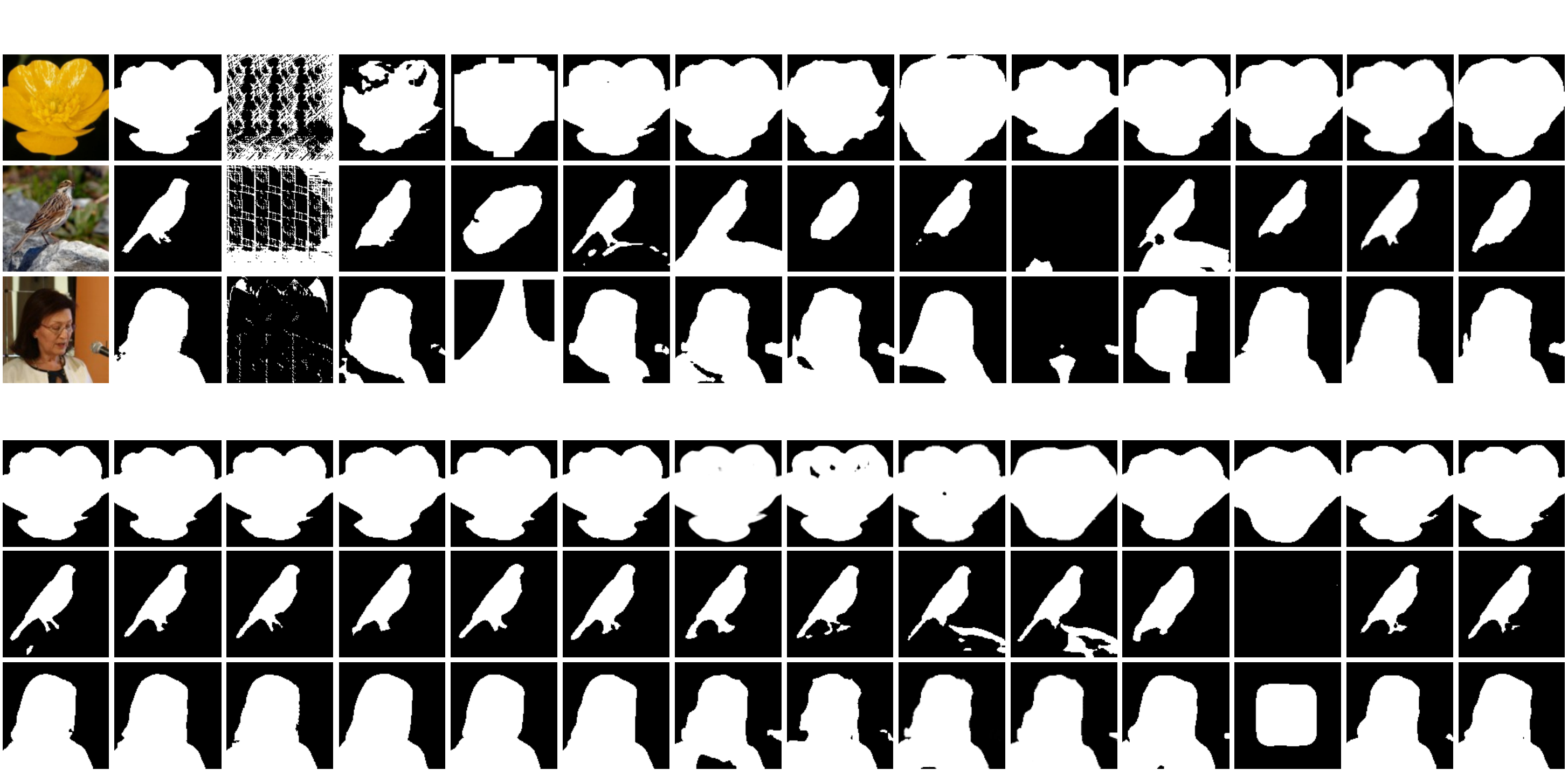}
\footnotesize{
\put(2.0,46.8){Input}
\put(9.1,46.8){GT}
\put(16.0,46.8){CAC}
\put(23.1,46.8){ReDO}
\put(29,46.8){SG-One}
\put(36.8,46.8){PANet}
\put(44.0,46.8){SPNet}
\put(51.0,46.8){CANet}
\put(58.0,46.8){ALSSS}
\put(65.5,46.8){USSS}
\put(70.5,46.8){Trans.(10)}
\put(78.2,46.8){Trans.(100)}
\put(87,48.5){Gated-}
\put(87,46.8){SCNN}
\put(95.0,46.8){BFP}

\put(2.1,21.8){Unet}
\put(9.1,21.8){FPN}
\put(15.0,21.8){LinkNet}
\put(22.5,21.8){PSPNet}
\put(30,21.8){PAN}
\put(34.5,21.8){DeepLabV3+}
\put(44.9,21.8){$T_{self}^{-}$}
\put(51.5,21.8){$T_{pseu}^{-}$}
\put(57.8,21.8){$T_{innner}^{-}$}
\put(65.5,21.8){$T_{outer}^{-}$}
\put(72.3,21.8){$T_{one D}$}
\put(79.9,21.8){\textbf{T}(0)}
\put(87.0,21.8){\textbf{T}(10)}
\put(94.5,21.8){\textbf{T}(all)}

}
\end{overpic}
\caption{The visual results of different methods on three datasets. Trans.($x$) denotes the DeepLabV3+ trained on $MixAll^{-}$ and finetuned with $x$ labeled target samples.
$\mathbi{T}_{self}^{-}$ and $\mathbi{T}_{pseu}^{-}$ denote the Trans-Net without boundary-aware self-supervision and pseudo triplet.
$\mathbi{T}_{inner}^{-}$ and $\mathbi{T}_{outer}^{-}$ denote the Trans-Net without inner discriminator and outer discriminator.
$\mathbi{T}_{one D}$ denotes the Trans-Net with only one discriminator.
$\mathbi{T}(x)$ denotes the Trans-Net with $x$ labeled samples of the target dataset.}
\label{fig:visualResults}
\end{figure*}
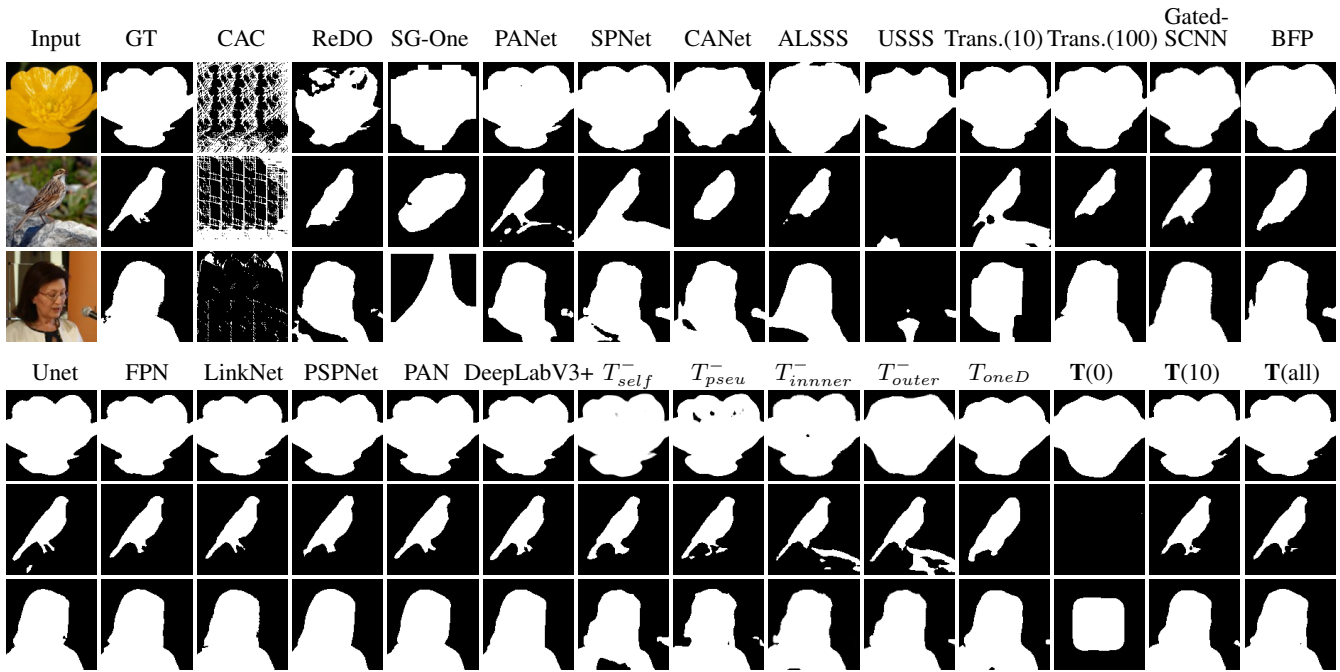


\subsection{Complete Algorithm}
To sum up, two boundary adversarial losses $\mathbf{\mathcal{L}^{out}_{adv}}$ and $\mathbf{\mathcal{L}^{in}_{adv}}$ are used for translating the visual boundary knowledge of the source dataset $\mathbf{S}^p$ into the segmentation network, which is designed for only segment samples from the target dataset $\mathbb{S}^q$.
The basic reconstruction loss $\mathbf{\mathcal{L}_{rec}}$ is adopted to supervise the segmentation on tens of labeled samples in target dataset $\mathbb{S}^q$. The self-supervised loss $\mathbf{\mathcal{L}_{sel}}$ is devised to strengthen the boundary consistency on target datasets $\mathbb{S}^q$.

During training, we alternatively optimize the segmentation network $\mathcal{F}_{\theta}$ and two boundary discriminators $ \mathcal{D}^o_{\phi},\mathcal{D}^i_{\varphi}$ using the randomly sampled samples from the  target dataset $\mathbb{S}^q$ and the source dataset $\mathbf{S}^p$, respectively. The complete algorithm is summarized in Algorithm~\ref{alg:alg1}. Once trained, the segmentation network concentrates on segmenting the foreground of the target category only.

\begin{table*}[!t]
\footnotesize
\caption{The performance comparison between SOTA methods and Trans-Net. The \emph{T(x)} denotes the Trans-Net with $x$ labeled samples of the target dataset. (All scores in $\%$). Trans.(x) denotes the DeepLabV3+ trained on $MixAll^{-}$ and finetuned with $x$ labeled target samples.
\textcolor[rgb]{0.80,0.00,0.00}{Red} and
\textcolor[rgb]{0.00,0.50,0.00}{Green} indicate the best and second-best performance. \textcolor[rgb]{0.00,0.00,1.00}{Blue} indicates the best performance among all non-fully supervised methods. Note that Flowers dataset contains only $753$ annotated samples.}
\label{comparing_SOTA}
\centering
\resizebox{\textwidth}{!}{
\begin{tabular}{ccccccccccccc}
\toprule
    & \multicolumn{4}{c}{\textbf{Birds} \textbf{Dataset}}
    & \multicolumn{4}{c}{\textbf{HumanMatting} \textbf{Dataset}}
    & \multicolumn{4}{c}{\textbf{Flowers} \textbf{Dataset}} \\
     \cmidrule(r){2-5}  \cmidrule(r){6-9} \cmidrule(r){10-13}
\textbf{Method}$\backslash$\textbf{Index}       &PA&MPA& MIoU& FWIoU   &PA&MPA& MIoU & FWIoU  &PA&MPA& MIoU  & FWIoU \\
    \cmidrule(r){1-1} \cmidrule(r){2-5}  \cmidrule(r){6-9} \cmidrule(r){10-13}

\textbf{CAC}        &41.90  &48.28 &26.42 & 27.06     &45.92 &47.14 &24.28 &23.72       &43.70 &36.96 &24.24 &35.34 \\

\textbf{ReDO}        &67.06  &50.00 &38.53 & 33.01     &51.44 &50.00 &35.72 & 36.49       &77.16 &70.00 &58.58 &43.82 \\
    \cmidrule(r){1-1} \cmidrule(r){2-5}  \cmidrule(r){6-9} \cmidrule(r){10-13}

\textbf{SG-One}        &83.45  &78.66 &61.43 & 74.68     &72.68 &72.46 &56.72 & 56.93       &86.47 &87.05 &74.73 &76.77 \\

\textbf{PANet}        &81.78  &66.94 &57.83 & 69.42     &73.89 &76.49 &60.54 & 63.29      &87.42 &68.43 &69.25 &69.94 \\
\textbf{SPNet} &85.21 &79.65 &76.92 &78.01 &77.82 &75.90 &60.42 &62.90 &87.90 &88.43 &79.21 &80.43\\
\textbf{CANet} &78.29 &73.49 &71.06 &69.84 &76.04 &73.80 &64.21 &63.90 &86.43 &84.96 &80.32 &79.16\\
    \cmidrule(r){1-1} \cmidrule(r){2-5}  \cmidrule(r){6-9} \cmidrule(r){10-13}
\textbf{ALSSS}        &76.95  &51.54 &39.48 & 66.09     &70.54 &76.26 &60.42 & 63.90       &87.43 &79.32 &85.21 &87.32 \\

\textbf{USSS}        &81.66  &49.64 &41.27 & 67.95     &77.25 &77.75 &62.40 & 62.28       &\textcolor[rgb]{0.00,0.00,1.00}{95.47} &\textcolor[rgb]{0.00,0.00,1.00}{95.15} &\textcolor[rgb]{0.00,0.00,1.00}{90.81} &\textcolor[rgb]{0.00,0.00,1.00}{91.36} \\
    \cmidrule(r){1-1} \cmidrule(r){2-5}  \cmidrule(r){6-9} \cmidrule(r){10-13}

\textbf{Trans.(10)}  &82.84  &79.83 &65.07 & 74.71     &85.32 &85.44 &75.71 & 75.72       &83.39 &82.69 &78.66 &76.39 \\
\textbf{Trans.(100)}    &91.26  &\textcolor[rgb]{0.00,0.00,1.00}{91.56} & \textcolor[rgb]{0.00,0.00,1.00}{83.23} & 84.06     &94.99 &95.00 &90.46 & 90.48       &93.90 &89.29 &81.33 &89.05 \\
    \cmidrule(r){1-1} \cmidrule(r){2-5}  \cmidrule(r){6-9} \cmidrule(r){10-13}
\textbf{Gated-SCNN} &83.23 &83.65 &57.93 &76.68 &81.18 &80.38 &67.66 &67.96 &86.81 &79.56 &71.18 &76.19\\
\textbf{BFP}        &80.47 &80.54 &58.99 &74.24 &79.32 &78.46 &65.34 &64.31 &84.28 &77.43 &68.29 &74.30\\
    \cdashline{1-13}[0.8pt/2pt]
\textbf{Unet}        &95.74  &91.88 &86.41  &92.06     &97.89 &97.88 &95.86 &95.87        &96.84 &96.39 &93.44 & 93.89 \\

\textbf{FPN}         &95.70  &92.86 &86.53  &92.06     &98.20 &98.19 &96.45 &96.46        &\textcolor[rgb]{0.00,0.50,0.00}{97.21} &\textcolor[rgb]{0.80,0.00,0.00}{97.16} &\textcolor[rgb]{0.00,0.50,0.00}{94.22} & \textcolor[rgb]{0.00,0.50,0.00}{94.59}\\

\textbf{LinkNet}     &95.50  &93.04 &86.03  &91.77     &97.42 &97.41 &94.97 &94.98        &\textcolor[rgb]{0.80,0.00,0.00}{97.25} &\textcolor[rgb]{0.00,0.50,0.00}{96.82} &\textcolor[rgb]{0.80,0.00,0.00}{94.26} & \textcolor[rgb]{0.80,0.00,0.00}{94.65}\\

\textbf{PSPNet}      &93.37  &87.01 &79.47  &87.97     &97.03 &97.02 &94.22 &94.23        &95.80 &95.77 &91.46 & 91.99\\

\textbf{PAN}         &95.86  &93.86 & 87.07 &92.38     &98.16 &98.15 &96.37 &96.38        &96.92 &96.71 &93.64 & 94.06\\

\textbf{DeepLabV3+}  &\textcolor[rgb]{0.00,0.50,0.00}{96.78}  &\textcolor[rgb]{0.80,0.00,0.00}{94.88} &\textcolor[rgb]{0.00,0.50,0.00}{89.62}  &\textcolor[rgb]{0.00,0.50,0.00}{93.95}     &\textcolor[rgb]{0.00,0.50,0.00}{98.28} &\textcolor[rgb]{0.00,0.50,0.00}{98.28} &\textcolor[rgb]{0.00,0.50,0.00}{96.62}& \textcolor[rgb]{0.00,0.50,0.00}{96.63}        &97.01 &96.65 &93.80 & 94.23\\
     \cmidrule(r){1-1} \cmidrule(r){2-5}  \cmidrule(r){6-9} \cmidrule(r){10-13}
\emph{\textbf{T}(0)}   &70.87  &56.19 &42.67 & 60.15     &66.41 &66.97 &48.90 &48.73        &83.04 &78.47 &67.78 & 70.56\\
\emph{\textbf{T}(10)}   &\textcolor[rgb]{0.00,0.00,1.00}{92.95}  &90.35 &79.58 & \textcolor[rgb]{0.00,0.00,1.00}{87.66}     &\textcolor[rgb]{0.00,0.00,1.00}{96.24} &\textcolor[rgb]{0.00,0.00,1.00}{96.29} &\textcolor[rgb]{0.00,0.00,1.00}{92.74} & \textcolor[rgb]{0.00,0.00,1.00}{92.75}       &93.20 &94.20 &88.70 & 89.30\\
\emph{\textbf{T}(all)}  &\textcolor[rgb]{0.80,0.00,0.00}{96.88}  &\textcolor[rgb]{0.00,0.50,0.00}{94.73} &\textcolor[rgb]{0.80,0.00,0.00}{89.76} & \textcolor[rgb]{0.80,0.00,0.00}{94.09}     &\textcolor[rgb]{0.80,0.00,0.00}{98.32} &\textcolor[rgb]{0.80,0.00,0.00}{98.32} &\textcolor[rgb]{0.80,0.00,0.00}{96.68} & \textcolor[rgb]{0.80,0.00,0.00}{96.68}        &95.92 &96.44 &91.77 & 92.23\\
\bottomrule
\end{tabular}}
\vspace{-1.em}
\end{table*}

\section{Experiments}

\textbf{Dataset.} The datasets we adopted contain single category datasets: Birds~\citep{Birds_datasets}, Flowers~\citep{nilsback2007delving}, and HumanMatting~\citep{Matting_human_datasets} and mixed category datasets: THUR15K~\citep{ChengGroupSaliency}, MSRA10K and MSRA-B~\citep{cheng2011global,hou2017deeply}, CSSD~\citep{yan2013hierarchical}, ECSSD~\citep{shi2016hierarchical}, DUT-OMRON~\citep{DUT_OMRON_datasets}, PASCAL-Context~\citep{mottaghi2014the}, HKU-IS~\citep{li2016visual}, SOD~\citep{movahedi2010design}, SIP1K~\citep{fan2019rethinking}.
The Birds, Flowers, HumanMatting and THUR15K are set as the target datasets in the experiments.  The Birds, Flowers, and HumanMatting contain $(11,788)$, $(8,189)$, $(34,427)$ samples, respectively.
The THUR15K contains $5$ categories and $15000$ samples.
The Flowers contains manually annotated parts ($753$ accurate masks) and algorithm~\citep{nilsback2007delving} pre-segmented parts ($8189$ rough masks).
In this paper, we adopt the manually labeled part. The experiments on the algorithm pre-segmented parts are given in the supplementary material. Except THUR15K, all the above mixed datasets are merged into the $MixAll$ dataset, where some of the mislabeled samples are deleted. The final $MixAll$ dataset contains $23,500$ samples.  When translating the visual boundary knowledge into the target category, the corresponding samples of the target category will be removed from $MixAll$, which generates the  $MixAll^{-}$ dataset. The data enhancement strategies for a few labeled dataset and affine transformation strategies are given in the supplementary materials.

\textbf{Network architecture.} In the paper, the segmentation network we adopted is the DeeplabV3+ (backbone: resnet50 )~\citep{chen2017rethinking}. Some popular network architectures including Unet~\citep{ronneberger2015u-net:}, FPN~\citep{lin2017feature}, Linknet~\citep{chaurasia2017linknet:}, PSPNet~\citep{zhao2017pyramid}, PAN~\citep{li2018pyramid} are also tested in our framework. Two boundary discriminators have the same encoder architecture, which is given in the supplementary materials.

\textbf{Parameter setting.} The parameters are set as follows: $\xi=1, \tau =1,\eta =1, \lambda = 10$, $n_{critic}=5$, the batch size $K =64$, Adam hyperparameters for two discriminators $\alpha =0.0001,\beta_{1}=0,\beta_{2}=0.9$. The learning rate for the segmentation network and two discriminators are all taken to be  $1e^{-4}$. The disk strel of radius $r$ is randomly sampled integer between $11$ and $55$.

\textbf{Metric.} The metrics we adopted include Pixel Accuracy (PA), Mean Pixel Accuracy (MPA), Mean Intersection over Union (MIoU), and Frequency Weighted Intersection over Union (FWIoU).

More details of datasets, network, parameters are given in the \emph{supplementary materials}.

\subsection{Comparing with SOTA Methods}

In this section, the proposed method is compared with the SOTA methods,
including \emph{unsupervised methods} :ReDO~\citep{chen2019unsupervised} and CAC~\citep{hsu2018co-attention},
\emph{few-shot methods}: SG-One~\citep{zhang2018sg-one:}, PANet~\citep{wang2019panet:}, SPNet~\citep{xian2019semantic} and CANet~\citep{zhang2019canet:},
\emph{weakly-/semi-supervised methods}: USSS~\citep{kalluri2019universal} and ALSSS~\citep{hung2018adversarial},
and \emph{fully supervised methods}: Unet~\citep{ronneberger2015u-net:}, FPN~\citep{lin2017feature}, LinkNet~\citep{chaurasia2017linknet:}, PSPNet~\citep{zhao2017pyramid}, PAN~\citep{li2018pyramid} and DeeplabV3+~\citep{chen2017rethinking} on four datasets.
Meanwhile, The Trans-Net is also compared with two boundary-aware methods: Gated-SCNN~\citep{takikawa2019gated-scnn} and BFP~\citep{ding2019boundary-aware}.
For the semi-supervised methods (USSS and ALSSS) and boundary-aware methods (Gated-SCNN and BFP), ten labeled samples are provided.
For $\mathbi{T}(x)$, the visual boundary knowledge is translated from the $MixAll^{-}$, which does not contain  samples from the target category.
Fig.~\ref{fig:visualResults} and Table~\ref{comparing_SOTA} show the quantitative and qualitative results, where we can see that most scores of proposed $\mathbi{T}(10)$ achieve the state-of-the-art results on par with existing non-fully supervised methods on Birds and HumanMatting datasets.
Note that Flowers~\citep{nilsback2007delving} dataset contains only $753$ manually annotated samples, which leads to the inconsistent scores with Birds and HumanMatting.
The most likely reason is overfitting. More experiments on the algorithm pre-segmented Flowers ($8189$ samples) are given in the supplementary material.
Moreover, with only $10$ labeled samples, $\mathbi{T}(10)$ can achieve better results than some fully supervised methods and close results on par with the best fully supervised method. Meanwhile, with all the labeled samples of the target dataset, the $\mathbi{T}(all)$ achieves almost all the highest scores.
In sum, such experiments demonstrate the practicability of the BKT-Task and the proposed Trans-Net.
More visual results and experiment results on THUR15K and algorithm pre-segmented Flowers are given in the supplementary materials.

\begin{table*}[!t]
\footnotesize
\caption{The translation results between different dataset setting.  `$\mathbb{S}^q \leftarrow \mathbf{S}^p$' denotes translating knowledge of source dataset $\mathbf{S}^p$ into the segmentation network for the target dataset $\mathbb{S}^q$ (All scores in $\%$).}
\label{task_on_datasets}
\centering
\resizebox{\textwidth}{!}{
\begin{tabular}{cccccc}
\toprule
\textbf{Index}$\backslash$\textbf{Task}
& \textbf{Humans} $\leftarrow$ \textbf{Birds}
& \textbf{Flowers} $\leftarrow$ \textbf{Birds}
& \textbf{(Birds, Flowers)} $\leftarrow$ \textbf{Humans}
& \textbf{(Humans, Birds, Flowers)} $\leftarrow$ \textbf{MixAll}$^{-}$ \\
\cmidrule(r){1-1}     \cmidrule(r){2-2}  \cmidrule(r){3-3}   \cmidrule(r){4-4}  \cmidrule(r){5-5}
PA                     &95.02            &92.70            &(92.87, 81.32)    & (94.82, 76.08, 86.10)  \\
MPA             &94.54            &94.00              &( 91.75, 78.78)    & (94.76, 84.15, 88.80)  \\
MIoU                   &88.86            &86.10             &(79.83, 66.38)    & (90.14, 56.33, 75.42)  \\
FWIoU                  &90.64            &86.70             &(87.65, 68.53)    & (90.16, 66.37, 76.01)  \\
\bottomrule
\end{tabular}
}
\end{table*}

\begin{table*}[!t]
\footnotesize 
\caption{The ablation study result of Trans-Net.
$\mathbi{T}_{self}^{-}$, $\mathbi{T}_{pseu}^{-}$ $\mathbi{T}_{inner}^{-}$ and $\mathbi{T}_{outer}^{-}$
denote the Trans-Net without boundary-aware self-supervision, pseudo triplet, inner discriminator and outer discriminator, respectively.
$\mathbi{T}_{one D}$ denotes the Trans-Net with only one discriminator.
$\mathbi{T}(x)$ denotes the Trans-Net with $x$ labeled samples of the target dataset. }
\label{ablation_table}
\centering
\resizebox{\textwidth}{!}{
\begin{tabular}{ccccccccccccc}
\toprule
\textbf{Index}$\backslash$\textbf{Ablation}
& $\mathbi{T}_{self}^{-}$
& $\mathbi{T}_{pseu}^{-}$
& $\mathbi{T}_{inner}^{-}$
& $\mathbi{T}_{outer}^{-}$
& $\mathbi{T}_{one D}$
& $\mathbi{T}(0)$
& $\mathbi{T}(5)$
& $\mathbi{T}(10)$
& $\mathbi{T}(20)$
& $\mathbi{T}(50)$
& $\mathbi{T}(100)$
& $\mathbi{T}(all)$ \\
\cmidrule(r){1-1} \cmidrule(r){2-3} \cmidrule(r){4-6} \cmidrule(r){7-13}
PA     &96.02  &95.98  & 93.08  & 95.28  &90.65  &66.41  &91.08  &96.24  &96.30  &96.91  &96.96  &98.32   \\
MPA    &95.94  &96.10  & 93.08  & 95.29 &90.54  &66.97  &91.25  &96.29  &96.31  &96.88  &96.95  &98.32   \\
MIoU   &92.12  &92.39  & 87.06  & 90.98 &82.85  &48.90  &83.62  &92.74  &92.87  &94.00  &94.09  &96.68   \\
FWIoU  &92.23  &91.70  &87.07   & 90.99 &82.89  &48.73  &83.62  &92.75  &92.88  &94.02  &94.10  &96.68   \\
\bottomrule
\end{tabular}
}
\end{table*}

\subsection{Knowledge Translation between Different Dataset}

To verify the robustness of the BKT-Task and the Trans-Net, the knowledge translation experiments between different datasets are provided. For all the tasks in Table~\ref{task_on_datasets}, only $10$ labeled samples of the target dataset are used in the Trans-Net.
We can see that all the translation tasks achieve satisfactory results.
Even for the translation task between very different categories (\{Humans $\leftarrow$ Birds\} and \{Flowers $\leftarrow$ Birds\}), Trans-Net still can achieve satisfactory results, which verifies the high extensibility and practicability of the BKT-Task and the Trans-Net.
For the scores on Humans, $\mathbi{T}(10)$ \{Humans $\leftarrow$ MixAll$^{-}$\} (Table~\ref{comparing_SOTA}) achieves better performance than \{(Humans, Birds, Flowers) $\leftarrow$ MixAll$^{-}$\} (Table~\ref{task_on_datasets}),
which confirms the assumption that Birds and Flowers will embezzle the segmentation capacity for Humans.
What's more, \{Humans $\leftarrow$ MixAll$^{-}$\} and \{Flowers $\leftarrow$ MixAll$^{-}$\} (Table~\ref{comparing_SOTA}) achieve higher scores than \{Humans $\leftarrow$ Birds\} and \{Flowers $\leftarrow$ Birds\}, which indicates that translating knowledge from the mixed category dataset is more proper for the BKT-Task.

Note that some scores of Birds in \{(Birds, Flowers) $\leftarrow$ MixAll$^{-}$\} are higher than
\{Birds $\leftarrow$ MixAll$^{-}$\}. One of the reasons is that the Humans dataset is a human matting dataset with more accurate segmentation GT than the MixAll$^{-}$ dataset, even we abandon some samples that are wicked GT in MixAll$^{-}$ dataset.
It also verifies that accurate boundary annotations are more proper for the BKT-Task.
The visual results of knowledge translation tasks between different datasets are given in the supplementary materials.

\subsection{Ablation Study}

To verify the effectiveness of the Trans-Net's components, we do ablation study on the boundary-aware self-supervised strategy, the pseudo triplet, two boundary discriminators, and the different numbers of labeled samples.
In this section, the knowledge translation task is set as \{Humans $\leftarrow$ MixAll$^{-}$\}.
For $\mathbi{T}_{self}^{-}$, $\mathbi{T}_{pseu}^{-}$, $\mathbi{T}_{inner}^{-}$, $\mathbi{T}_{inner}^{-}$ and $\mathbi{T}_{oneD}$, there are ten labeled samples of the Humans dataset. From Table~\ref{ablation_table}, we can see that Trans-Net achieves higher scores than $\mathbi{T}_{self}^{-}$ and $\mathbi{T}_{pseu}^{-}$, which demonstrates the effectiveness of the self-supervised strategy and the pseudo triplet.
Fig.~\ref{fig:visualResults} shows that the result of $\mathbi{T}_{self}^{-}$ has some holes. By contrast, the result of Trans-Net is accurate, which indicates that the boundary-aware self-supervised strategy is beneficial for eliminating the incorrect holes. For the $\mathbi{T}_{oneD}$, the inputs of two discriminators will be concatenated then be input into one discriminator. The scores of $\mathbi{T}_{oneD}$ have dropped by about $10\%$ comparing with Trans-Net on all indexes,  which demonstrates that the two-discriminator framework is useful for improving the segmentation performance.
Meanwhile, $\mathbi{T}(10)$ achieves
about $4\%$ increase on the scores of $\mathbi{T}_{inner}^{-}$ and about $2\%$ increase on the scores of $\mathbi{T}_{outer}^{-}$,
which demonstrates the effectiveness of inner and outer boundary discriminators.
For the different numbers of labeled samples, we find that $10$-labeled-samples is a critical cut-off point, which can supply relatively sufficient guidance. An object usually contains multiple components, which have different edges. Without any guidance of labeled samples, the segmentation network will not be aware of
the desired boundary of the object. Therefore, the $\mathbi{T}(0)$ achieves the worst performance.
With all labeled samples, $\mathbi{T}(all)$ achieves SOTA results on par with existing fully supervised methods.
More visual results of the ablation study are given in the supplementary materials.


\section{Conclusion}

In this paper, we study a new Boundary Knowledge Translation Task (BKT-Task). This is inspired by the fact that humans can perfectly segment the object from an image according to the boundary information without knowing the object's category.
The goal of BKT-Task is to translate the visual boundary knowledge of source datasets into the segmentation network for new categories
in a least effort and dependable way.

Based on the proposed BKT-Task, we introduce the Translation Segmentation Network (Trans-Net) for segmenting the foreground from the background. The Trans-Net contains a segmentation network and two boundary discriminators, which are devised for translating visual boundary knowledge from the source dataset to the target one. Meanwhile, boundary-aware self-supervision and pseudo triplet are devised to enhance boundary consistency and help the two discriminators distinguish boundary, respectively. Exhaustive experiments verify the promising generalization and practicability of the BKT-Task. Furthermore, with only tens of labeled sample of the target dataset, the Trans-Net achieves results on par with fully supervised methods. In the future, we will generalize the knowledge translation task into other applications, such as image matting and image classification. Meanwhile, we will extend the Trans-Net into more general multiple object image segmentation.

\section{ Acknowledgments}
This work was supported by National Natural Science Foundation of China (No.62002318), Key Research and Development Program of Zhejiang Province (2020C01023), Zhejiang Provincial Science and Technology Project for Public Welfare (LGF21F020020), Programs Supported by Ningbo Natural Science Foundation (202003N4318), and Major Scientifc Research Project of Zhejiang Lab (No. 2019KD0AC01).

\section*{Potential Ethical Impact }

We discuss the potential positive and negative impact of the proposed work here.  Positive: the proposed method can be applied to image foreground segmentation, and also  be extended to other general image segmentation in the future, which is beneficial for image editing applications.
Moreover, the proposed Boundary Knowledge
Translation Task (BKT-Task) will
potentially
change the conventional way that the category-aware label supervises the model learning of corresponding category-aware knowledge.
With vast public annotated datasets, training a segmentation network for the new category requires only tens of labeled samples of the new category. The BKT-Task will dramatically reduce the requirement for the labeled samples of new category, while the trained model  achieves gratifying performances. Negative: the research might be adopted to generate  fake images, which can be used for malicious purposes.

\bibliography{MYRE}

\begin{thebibliography}{58}
\providecommand{\natexlab}[1]{#1}
\providecommand{\url}[1]{\texttt{#1}}
\providecommand{\urlprefix}{URL }
\expandafter\ifx\csname urlstyle\endcsname\relax
  \providecommand{\doi}[1]{doi:\discretionary{}{}{}#1}\else
  \providecommand{\doi}{doi:\discretionary{}{}{}\begingroup
  \urlstyle{rm}\Url}\fi

\bibitem[{Arbelle and Raviv(2018)}]{arbelle2018microscopy}
Arbelle, A.; and Raviv, T.~R. 2018.
\newblock Microscopy Cell Segmentation via Adversarial Neural Networks.
\newblock \emph{ISBI} 645--648.

\bibitem[{Bertasius, Shi, and Torresani(2016)}]{bertasius2016semantic}
Bertasius, G.; Shi, J.; and Torresani, L. 2016.
\newblock Semantic Segmentation with Boundary Neural Fields.
\newblock \emph{CVPR} 3602--3610.

\bibitem[{Catherine et~al.(2011)Catherine, Steve, Peter, Pietro, and
  Serge}]{Birds_datasets}
Catherine, W.; Steve, B.; Peter, W.; Pietro, P.; and Serge, B. 2011.
\newblock The Caltech-UCSD Birds-200-2011 Dataset.
\newblock Technical Report CNS-TR-2011-001, California Institute of Technology.

\bibitem[{Chaurasia and Culurciello(2017)}]{chaurasia2017linknet:}
Chaurasia, A.; and Culurciello, E. 2017.
\newblock LinkNet: Exploiting encoder representations for efficient semantic
  segmentation.
\newblock \emph{VCIP} 1--4.

\bibitem[{Chen et~al.(2016)Chen, Barron, Papandreou, Murphy, and
  Yuille}]{chen2016semantic}
Chen, L.; Barron, J.~T.; Papandreou, G.; Murphy, K.; and Yuille, A.~L. 2016.
\newblock Semantic Image Segmentation with Task-Specific Edge Detection Using
  CNNs and a Discriminatively Trained Domain Transform.
\newblock \emph{CVPR} 4545--4554.

\bibitem[{Chen et~al.(2017)Chen, Papandreou, Schroff, and
  Adam}]{chen2017rethinking}
Chen, L.; Papandreou, G.; Schroff, F.; and Adam, H. 2017.
\newblock Rethinking Atrous Convolution for Semantic Image Segmentation.
\newblock \emph{arXiv} .

\bibitem[{Chen, Artieres, and Denoyer(2019)}]{chen2019unsupervised}
Chen, M.; Artieres, T.; and Denoyer, L. 2019.
\newblock Unsupervised Object Segmentation by Redrawing.
\newblock \emph{NeurIPS} 12726--12737.

\bibitem[{Cheng et~al.(2017)Cheng, Meng, Xiang, and Pan}]{heng2017fusionNet:}
Cheng, D.; Meng, G.; Xiang, S.; and Pan, C. 2017.
\newblock FusionNet: Edge Aware Deep Convolutional Networks for Semantic
  Segmentation of Remote Sensing Harbor Images.
\newblock \emph{IEEE JSTAEORS} 10(12): 5769--5783.

\bibitem[{Cheng et~al.(2011)Cheng, Zhang, Mitra, Huang, and
  Hu}]{cheng2011global}
Cheng, M.; Zhang, G.; Mitra, N.~J.; Huang, X.; and Hu, S. 2011.
\newblock Global contrast based salient region detection.
\newblock \emph{CVPR} 37(3): 409--416.

\bibitem[{Cheng et~al.(2014)Cheng, Mitra, Huang, and Hu}]{ChengGroupSaliency}
Cheng, M.-M.; Mitra, N.; Huang, X.; and Hu, S.-M. 2014.
\newblock SalientShape: group saliency in image collections.
\newblock \emph{The Visual Computer} 30(4): 443--453.

\bibitem[{Company(2019)}]{Matting_human_datasets}
Company, A. 2019.
\newblock Matting human datasets.
\newblock \url{https://github.com/aisegmentcn/matting_human_datasets}.

\bibitem[{Dai et~al.(2019)Dai, Mo, Angelini, Guo, and Bai}]{Dai2019Transfer}
Dai, C.; Mo, Y.; Angelini, E.; Guo, Y.; and Bai, W. 2019.
\newblock Transfer Learning from Partial Annotations for Whole Brain
  Segmentation.
\newblock \emph{MICCAI Workshop} 199--206.

\bibitem[{Ding et~al.(2019)Ding, Jiang, Liu, Thalmann, and
  Wang}]{ding2019boundary-aware}
Ding, H.; Jiang, X.; Liu, A.~Q.; Thalmann, N.~M.; and Wang, G. 2019.
\newblock Boundary-Aware Feature Propagation for Scene Segmentation.
\newblock \emph{ICCV} 6819--6829.

\bibitem[{Fan et~al.(2019)Fan, Lin, Zhao, Liu, Zhang, Hou, Zhu, and
  Cheng}]{fan2019rethinking}
Fan, D.; Lin, Z.; Zhao, J.; Liu, Y.; Zhang, Z.; Hou, Q.; Zhu, M.; and Cheng, M.
  2019.
\newblock Rethinking RGB-D Salient Object Detection: Models, Datasets, and
  Large-Scale Benchmarks.
\newblock \emph{arXiv} .

\bibitem[{Gulrajani et~al.(2017)Gulrajani, Ahmed, Arjovsky, Dumoulin, and
  Courville}]{gulrajani2017improved}
Gulrajani, I.; Ahmed, F.; Arjovsky, M.; Dumoulin, V.; and Courville, A. 2017.
\newblock Improved training of wasserstein GANs.
\newblock \emph{NeurIPS} 5769--5779.

\bibitem[{Han and Yin(2017)}]{han2017transferring}
Han, L.; and Yin, Z. 2017.
\newblock Transferring Microscopy Image Modalities with Conditional Generative
  Adversarial Networks.
\newblock \emph{CVPR} 2017: 851--859.

\bibitem[{Hayder, He, and Salzmann(2017)}]{hayder2017boundary-aware}
Hayder, Z.; He, X.; and Salzmann, M. 2017.
\newblock Boundary-Aware Instance Segmentation.
\newblock \emph{CVPR} 587--595.

\bibitem[{Hong et~al.(2017)Hong, Yeo, Kwak, Lee, and Han}]{hong2017weakly}
Hong, S.; Yeo, D.; Kwak, S.; Lee, H.; and Han, B. 2017.
\newblock Weakly Supervised Semantic Segmentation Using Web-Crawled Videos.
\newblock \emph{CVPR} 2224--2232.

\bibitem[{Hou et~al.(2017)Hou, Cheng, Hu, Borji, Tu, and Torr}]{hou2017deeply}
Hou, Q.; Cheng, M.; Hu, X.; Borji, A.; Tu, Z.; and Torr, P. H.~S. 2017.
\newblock Deeply Supervised Salient Object Detection with Short Connections.
\newblock \emph{CVPR} 5300--5309.

\bibitem[{Hsu, Lin, and Chuang(2018)}]{hsu2018co-attention}
Hsu, K.; Lin, Y.; and Chuang, Y. 2018.
\newblock Co-attention CNNs for Unsupervised Object Co-segmentation.
\newblock \emph{IJCAI} 748--756.

\bibitem[{Hung et~al.(2018)Hung, Tsai, Liou, Lin, and
  Yang}]{hung2018adversarial}
Hung, W.; Tsai, Y.; Liou, Y.; Lin, Y.; and Yang, M. 2018.
\newblock Adversarial Learning for Semi-Supervised Semantic Segmentation.
\newblock \emph{BMVC} 65.

\bibitem[{Kalluri et~al.(2019)Kalluri, Varma, Chandraker, and
  Jawahar}]{kalluri2019universal}
Kalluri, T.; Varma, G.; Chandraker, M.; and Jawahar, C.~V. 2019.
\newblock Universal Semi-Supervised Semantic Segmentation.
\newblock \emph{ICCV} 5259--5270.

\bibitem[{Khoreva et~al.(2017)Khoreva, Benenson, Hosang, Hein, and
  Schiele}]{khoreva2017simple}
Khoreva, A.; Benenson, R.; Hosang, J.; Hein, M.; and Schiele, B. 2017.
\newblock Simple Does It: Weakly Supervised Instance and Semantic Segmentation.
\newblock \emph{CVPR} 1665--1674.

\bibitem[{Li and Yu(2016)}]{li2016visual}
Li, G.; and Yu, Y. 2016.
\newblock Visual Saliency Detection Based on Multiscale Deep CNN Features.
\newblock \emph{IEEE TIP} 5012--5024.

\bibitem[{Li et~al.(2018)Li, Xiong, An, and Wang}]{li2018pyramid}
Li, H.; Xiong, P.; An, J.; and Wang, L. 2018.
\newblock Pyramid Attention Network for Semantic Segmentation.
\newblock \emph{BMVC} 285.

\bibitem[{Li, Arnab, and Torr(2018)}]{li2018weakly-}
Li, Q.; Arnab, A.; and Torr, P. H.~S. 2018.
\newblock Weakly- and Semi-Supervised Panoptic Segmentation.
\newblock \emph{ECCV} 102--118.

\bibitem[{Lin et~al.(2017)Lin, Dollar, Girshick, He, Hariharan, and
  Belongie}]{lin2017feature}
Lin, T.; Dollar, P.; Girshick, R.; He, K.; Hariharan, B.; and Belongie, S.
  2017.
\newblock Feature Pyramid Networks for Object Detection.
\newblock \emph{CVPR} 936--944.

\bibitem[{Luc et~al.(2016)Luc, Couprie, Chintala, and
  Verbeek}]{luc2016semantic}
Luc, P.; Couprie, C.; Chintala, S.; and Verbeek, J. 2016.
\newblock Semantic Segmentation using Adversarial Networks.
\newblock \emph{NIPS} .

\bibitem[{Minaee et~al.(2020)Minaee, Boykov, Porikli, Plaza, Kehtarnavaz, and
  Terzopoulos}]{minaee2020image}
Minaee, S.; Boykov, Y.; Porikli, F.; Plaza, A.; Kehtarnavaz, N.; and
  Terzopoulos, D. 2020.
\newblock Image Segmentation Using Deep Learning: A Survey.
\newblock \emph{arXiv} .

\bibitem[{Mottaghi et~al.(2014)Mottaghi, Chen, Liu, Cho, Lee, Fidler, Urtasun,
  and Yuille}]{mottaghi2014the}
Mottaghi, R.; Chen, X.; Liu, X.; Cho, N.; Lee, S.; Fidler, S.; Urtasun, R.; and
  Yuille, A.~L. 2014.
\newblock The Role of Context for Object Detection and Semantic Segmentation in
  the Wild.
\newblock \emph{CVPR} 891--898.

\bibitem[{Movahedi and Elder(2010)}]{movahedi2010design}
Movahedi, V.; and Elder, J.~H. 2010.
\newblock Design and perceptual validation of performance measures for salient
  object segmentation.
\newblock \emph{CVPR} 49--56.

\bibitem[{Nilsback and Zisserman(2007)}]{nilsback2007delving}
Nilsback, M.~E.; and Zisserman, A. 2007.
\newblock Delving into the whorl of flower segmentation.
\newblock \emph{BMVC} 1--10.

\bibitem[{Ostyakov et~al.(2018)Ostyakov, Suvorov, Logacheva, Khomenko, and
  Nikolenko}]{ostyakov2018seigan:}
Ostyakov, P.; Suvorov, R.; Logacheva, E.; Khomenko, O.; and Nikolenko, S.~I.
  2018.
\newblock SEIGAN: Towards Compositional Image Generation by Simultaneously
  Learning to Segment, Enhance, and Inpaint.
\newblock \emph{arXiv} .

\bibitem[{Papandreou et~al.(2015)Papandreou, Chen, Murphy, and
  Yuille}]{papandreou2015weakly-and}
Papandreou, G.; Chen, L.; Murphy, K.; and Yuille, A.~L. 2015.
\newblock Weakly-and Semi-Supervised Learning of a Deep Convolutional Network
  for Semantic Image Segmentation.
\newblock \emph{ICCV} 1742--1750.

\bibitem[{Peng et~al.(2017)Peng, Zhang, Yu, Luo, and Sun}]{peng2017large}
Peng, C.; Zhang, X.; Yu, G.; Luo, G.; and Sun, J. 2017.
\newblock Large Kernel Matters ¡ª Improve Semantic Segmentation by Global
  Convolutional Network.
\newblock \emph{CVPR} 1743--1751.

\bibitem[{Qin et~al.(2019)Qin, Zhang, Huang, Gao, Dehghan, and
  Jagersand}]{qin2019basnet:}
Qin, X.; Zhang, Z.; Huang, C.; Gao, C.; Dehghan, M.; and Jagersand, M. 2019.
\newblock BASNet: Boundary-Aware Salient Object Detection.
\newblock \emph{CVPR} 7479--7489.

\bibitem[{Remez, Huang, and Brown(2018)}]{remez2018learning}
Remez, T.; Huang, J.; and Brown, M. 2018.
\newblock Learning to Segment via Cut-and-Paste.
\newblock \emph{ECCV} 39--54.

\bibitem[{Ronneberger, Fischer, and Brox(2015)}]{ronneberger2015u-net:}
Ronneberger, O.; Fischer, P.; and Brox, T. 2015.
\newblock U-Net: Convolutional Networks for Biomedical Image Segmentation.
\newblock \emph{MICCAI} 234--241.

\bibitem[{Ruan, Tong, and Lu(2011)}]{DUT_OMRON_datasets}
Ruan, X.; Tong, N.; and Lu, H. 2011.
\newblock How far we away from a perfect visual saliency detection - DUT-OMRON:
  a new benchmark dataset.
\newblock FCV2014.

\bibitem[{Shaban et~al.(2017)Shaban, Bansal, Liu, Essa, and
  Boots}]{shaban2017one-shot}
Shaban, A.; Bansal, S.; Liu, Z.; Essa, I.; and Boots, B. 2017.
\newblock One-Shot Learning for Semantic Segmentation.
\newblock \emph{BMVC} .

\bibitem[{Shi et~al.(2016)Shi, Yan, Xu, and Jia}]{shi2016hierarchical}
Shi, J.; Yan, Q.; Xu, L.; and Jia, J. 2016.
\newblock Hierarchical Image Saliency Detection on Extended CSSD.
\newblock \emph{IEEE TPAMI} 38(4): 717--729.

\bibitem[{Siam, Oreshkin, and Jagersand(2019)}]{siam2019amp:}
Siam, M.; Oreshkin, B.~N.; and Jagersand, M. 2019.
\newblock AMP: Adaptive Masked Proxies for Few-Shot Segmentation.
\newblock \emph{ICCV} 5249--5258.

\bibitem[{Souly, Spampinato, and Shah(2017)}]{souly2017semi}
Souly, N.; Spampinato, C.; and Shah, M. 2017.
\newblock Semi Supervised Semantic Segmentation Using Generative Adversarial
  Network.
\newblock \emph{ICCV} 5689--5697.

\bibitem[{Sun et~al.(2019)Sun, Zhu, Wu, Huang, Shi, and Ma}]{sun2019not}
Sun, R.; Zhu, X.; Wu, C.; Huang, C.; Shi, J.; and Ma, L. 2019.
\newblock Not All Areas Are Equal: Transfer Learning for Semantic Segmentation
  via Hierarchical Region Selection.
\newblock \emph{CVPR} 4360--4369.

\bibitem[{Takikawa et~al.(2019{\natexlab{a}})Takikawa, Acuna, Jampani, and
  Fidler}]{takikawa2019gated-scnn:}
Takikawa, T.; Acuna, D.; Jampani, V.; and Fidler, S. 2019{\natexlab{a}}.
\newblock Gated-SCNN: Gated Shape CNNs for Semantic Segmentation.
\newblock \emph{arXiv} .

\bibitem[{Takikawa et~al.(2019{\natexlab{b}})Takikawa, Acuna, Jampani, and
  Fidler}]{takikawa2019gated-scnn}
Takikawa, T.; Acuna, D.; Jampani, V.; and Fidler, S. 2019{\natexlab{b}}.
\newblock Gated-SCNN: Gated Shape CNNs for Semantic Segmentation.
\newblock \emph{ICCV} 5229--5238.

\bibitem[{Vinyals et~al.(2016)Vinyals, Blundell, Lillicrap, Kavukcuoglu, and
  Wierstra}]{vinyals2016matching}
Vinyals, O.; Blundell, C.; Lillicrap, T.; Kavukcuoglu, K.; and Wierstra, D.
  2016.
\newblock Matching Networks for One Shot Learning.
\newblock \emph{NeurIPS} 3637--3645.

\bibitem[{Wang et~al.(2019{\natexlab{a}})Wang, Liew, Zou, Zhou, and
  Feng}]{wang2019panet:}
Wang, K.; Liew, J.~H.; Zou, Y.; Zhou, D.; and Feng, J. 2019{\natexlab{a}}.
\newblock PANet: Few-Shot Image Semantic Segmentation With Prototype Alignment.
\newblock \emph{ICCV} 9197--9206.

\bibitem[{Wang et~al.(2019{\natexlab{b}})Wang, Zhang, Kan, Shan, and
  Chen}]{wang2019self-supervised}
Wang, Y.; Zhang, J.; Kan, M.; Shan, S.; and Chen, X. 2019{\natexlab{b}}.
\newblock Self-supervised Scale Equivariant Network for Weakly Supervised
  Semantic Segmentation.
\newblock \emph{arXiv} .

\bibitem[{Wang, Simoncelli, and Bovik(2003)}]{wang2003multiscale}
Wang, Z.; Simoncelli, E.~P.; and Bovik, A.~C. 2003.
\newblock Multiscale Structural Similarity for Image Quality Assessment.
\newblock \emph{ACSSC} 2: 1398--1402.

\bibitem[{Xian et~al.(2019)Xian, Choudhury, He, Schiele, and
  Akata}]{xian2019semantic}
Xian, Y.; Choudhury, S.; He, Y.; Schiele, B.; and Akata, Z. 2019.
\newblock Semantic Projection Network for Zero- and Few-Label Semantic
  Segmentation.
\newblock \emph{CVPR} 8256--8265.

\bibitem[{Xue et~al.(2018)Xue, Xu, Zhang, Long, and Huang}]{xue2018segan:}
Xue, Y.; Xu, T.; Zhang, H.; Long, L.~R.; and Huang, X. 2018.
\newblock SegAN: Adversarial Network with Multi-scale L 1 Loss for Medical
  Image Segmentation.
\newblock \emph{Neuroinformatics} 16(3): 383--392.

\bibitem[{Yan et~al.(2013)Yan, Xu, Shi, and Jia}]{yan2013hierarchical}
Yan, Q.; Xu, L.; Shi, J.; and Jia, J. 2013.
\newblock Hierarchical Saliency Detection.
\newblock \emph{CVPR} 1155--1162.

\bibitem[{Yu et~al.(2018)Yu, Wang, Peng, Gao, Yu, and Sang}]{yu2018learning}
Yu, C.; Wang, J.; Peng, C.; Gao, C.; Yu, G.; and Sang, N. 2018.
\newblock Learning a Discriminative Feature Network for Semantic Segmentation.
\newblock \emph{CVPR} 1857--1866.

\bibitem[{Zhang et~al.(2019)Zhang, Lin, Liu, Yao, and Shen}]{zhang2019canet:}
Zhang, C.; Lin, G.; Liu, F.; Yao, R.; and Shen, C. 2019.
\newblock CANet: Class-Agnostic Segmentation Networks With Iterative Refinement
  and Attentive Few-Shot Learning.
\newblock \emph{CVPR} 5217--5226.

\bibitem[{Zhang et~al.(2017)Zhang, Tang, Lin, Li, and
  Yan}]{zhang2017global-residual}
Zhang, R.; Tang, S.; Lin, M.; Li, J.; and Yan, S. 2017.
\newblock Global-residual and local-boundary refinement networks for rectifying
  scene parsing predictions.
\newblock \emph{IJCAI} 3427--3433.

\bibitem[{Zhang et~al.(2018)Zhang, Wei, Yang, and Huang}]{zhang2018sg-one:}
Zhang, X.; Wei, Y.; Yang, Y.; and Huang, T.~S. 2018.
\newblock SG-One: Similarity Guidance Network for One-Shot Semantic
  Segmentation.
\newblock \emph{IEEE TCYB} .

\bibitem[{Zhao et~al.(2017)Zhao, Shi, Qi, Wang, and Jia}]{zhao2017pyramid}
Zhao, H.; Shi, J.; Qi, X.; Wang, X.; and Jia, J. 2017.
\newblock Pyramid Scene Parsing Network.
\newblock \emph{CVPR} 6230--6239.

\end{thebibliography}

\end{document}